\documentclass[a4paper,fleqn]{cas-dc}

\usepackage[authoryear]{natbib}
\usepackage{bbm}
\usepackage{textcomp}
\usepackage{subfigure}
\usepackage{multirow}
\usepackage{makecell}
\usepackage{arydshln} 
\usepackage{xcolor}
\usepackage[ruled,lined]{algorithm2e}

\def\tsc#1{\csdef{#1}{\textsc{\lowercase{#1}}\xspace}}
\tsc{WGM}
\tsc{QE}

\begin{document}
\let\WriteBookmarks\relax
\def\floatpagepagefraction{1}
\def\textpagefraction{.001}
\let\printorcid\relax

\shorttitle{}    

\shortauthors{Tianling Liu et al.}  

\title [mode = title]{CFCML: A Coarse-to-Fine Crossmodal Learning Framework For Disease Diagnosis Using Multimodal Images and Tabular Data}  



%
















\author[1]{Tianling Liu}
\ead{liu_dling@tju.edu.cn}

\author[2,3]{Hongying Liu}
\ead{hyliu2009@tju.edu.cn}

\author[1]{Fanhua Shang}
\ead{fhshang@tju.edu.cn}

\author[4]{Lequan Yu
\corref{correspondingauthor}}
\ead{lqyu@hku.hk}

\author[5,6]{Tong Han}
\ead{mrbold@163.com}

\author[1,2]{Liang Wan
\corref{correspondingauthor}}
\ead{lwan@tju.edu.cn}

\cortext[correspondingauthor]{Corresponding author: Lequan Yu, Liang Wan.}

\address[1]{College of Intelligence and Computing, Tianjin University, Tianjin 300350, China.}

\address[2]{Medical School of Tianjin University, Tianjin 300072, China.}

\address[3]{Peng Cheng Lab, Shenzhen 518055, China.}

\address[4]{ Department of Statistics and Actuarial Science, School of Computing and Data Science, The University of Hong Kong, Hong Kong.}

\address[5]{Department of Radiology, Tianjin Huanhu Hospital, Tianjin 300350, China.}

\address[6]{Tianjin Key Laboratory of Cerebral Vascular and Neurodegenerative Diseases, Tianjin 300350, China.}


\begin{abstract}
In clinical practice, crossmodal information including medical images and tabular data is essential for disease diagnosis. There exists a significant modality gap between these data types, which obstructs advancements in crossmodal diagnostic accuracy.
Most existing crossmodal learning (CML) methods primarily focus on exploring relationships among high-level encoder outputs, leading to the neglect of local information in images. Additionally, these methods often overlook the extraction of task-relevant information.
In this paper, we propose a novel coarse-to-fine crossmodal learning (CFCML) framework to progressively reduce the modality gap between multimodal images and tabular data, by thoroughly exploring inter-modal relationships.
At the coarse stage, we explore the relationships between multi-granularity features from various image encoder stages and tabular information, facilitating a preliminary reduction of the modality gap. 
At fine stage, we generate unimodal and crossmodal prototypes that incorporate class-aware information, and establish hierarchical anchor-based relationship mining (HRM) strategy to further diminish the modality gap and extract discriminative crossmodal information. This strategy utilize modality samples, unimodal prototypes, and crossmodal prototypes as anchors to develop contrastive learning approaches, effectively enhancing inter-class disparity while reducing intra-class disparity from multiple perspectives.
Experimental results indicate that our method outperforms the state-of-the-art (SOTA) methods, achieving improvements of 1.53\% and 0.91\% in AUC metrics on the MEN and Derm7pt datasets, respectively.
The code is available at https://github.com/IsDling/CFCML.

\end{abstract}





\maketitle

\section{Introduction}
\label{sec_introduction}
In clinical practice, physicians typically diagnose diseases by integrating various information sources~\citep{huang2020fusion}, such as medical images and clinical data. Medical images often provide critical insights into lesions and anatomical structures, enabling visual assessment for patients. In contrast, clinical data include additional essential information such as age, medical history, and details of lesion area. The integration of multimodal medical images and clinical data exemplifies crossmodal learning (CML), enabling a more comprehensive understanding of the patient's health status and facilitating more accurate diagnoses and treatment planning.

CML has garnered significant attentions from researchers due to its potential to provide more comprehensive and robust features~\citep{han2022trusted, li2023decoupled,xu2022remixformer,grzeszczyk2023tabattention}.
Some studies employ concatenation operations~\citep{xia2019multi,el2020multimodal, holste2021end} to fuse the crossmodal information, while other methods~\citep{han2022multimodal, han2022trusted} focus on exploring the uncertainty associated with each modality and generating fusion weights based on the learned uncertainty information. Nevertheless, these approaches often fail to fully explore the interrelationships between cross modalities and frequently overlook the inherent heterogeneity among them, which poses challenges to the learning process.
Many studies aim to address the modality gap through the feature disentanglement paradigm, which typically explores both modality-shared and modality-specific features~\citep{yang2022disentangled, li2023decoupled, wang2023shared}. Specifically, modality-shared features can reduce the modality gap by providing a unified representation derived from different modalities. In contrast, extracting modality-specific features allows a deeper understanding of the unique characteristics inherent to each modality, thereby facilitating the acquisition of comprehensive crossmodal information.
Furthermore, additional research~\citep{song2021cross,zhu2022multimodal,hu2025itcfn} focuses on reducing the modality gap using the cross-attention mechanism by exploring the correlations between modalities while obtaining complementary information.

However, in the context of medical multimodal images-tabular fusion, the lesion location information in the tabular data relies on the overall information from the images for accurate correspondence, typically captured effectively in the deeper stages of the encoder. Conversely, lesion content information corresponds precisely to the lesion regions in the images, which are effectively captured in the shallower stages. Therefore, aligning tabular data with images necessitates careful consideration of the relationship exploration between the multi-granularity features from the multimodal images and tabular data. To our knowledge, existing CML studies have primarily focused on high-level inter-modal relationship, which may hinder efforts to reduce the modality gap.
Furthermore, these studies often overlook the extraction of class-aware features that are essential for disease diagnostic during intermodal relationship exploration.
This oversight may lead to redundant information, ultimately affecting overall performance. Yan et al.~\citep{yan2024causality} proposed the Modality-Relevant Interactive Mining (MRIM) module, which introduces class information to mitigate the modality gap between crossmodal images.
Despite the success of MRIM, it primarily targets two image modalities, which have a smaller modality gap compared to that between images and tabular data. Additionally, the mining of crossmodal relationship is insufficient, making it challenging to adapt to application scenarios involving multimodal images and tabular data.


\begin{figure*}
    \centering
    \includegraphics[width=0.9\linewidth]{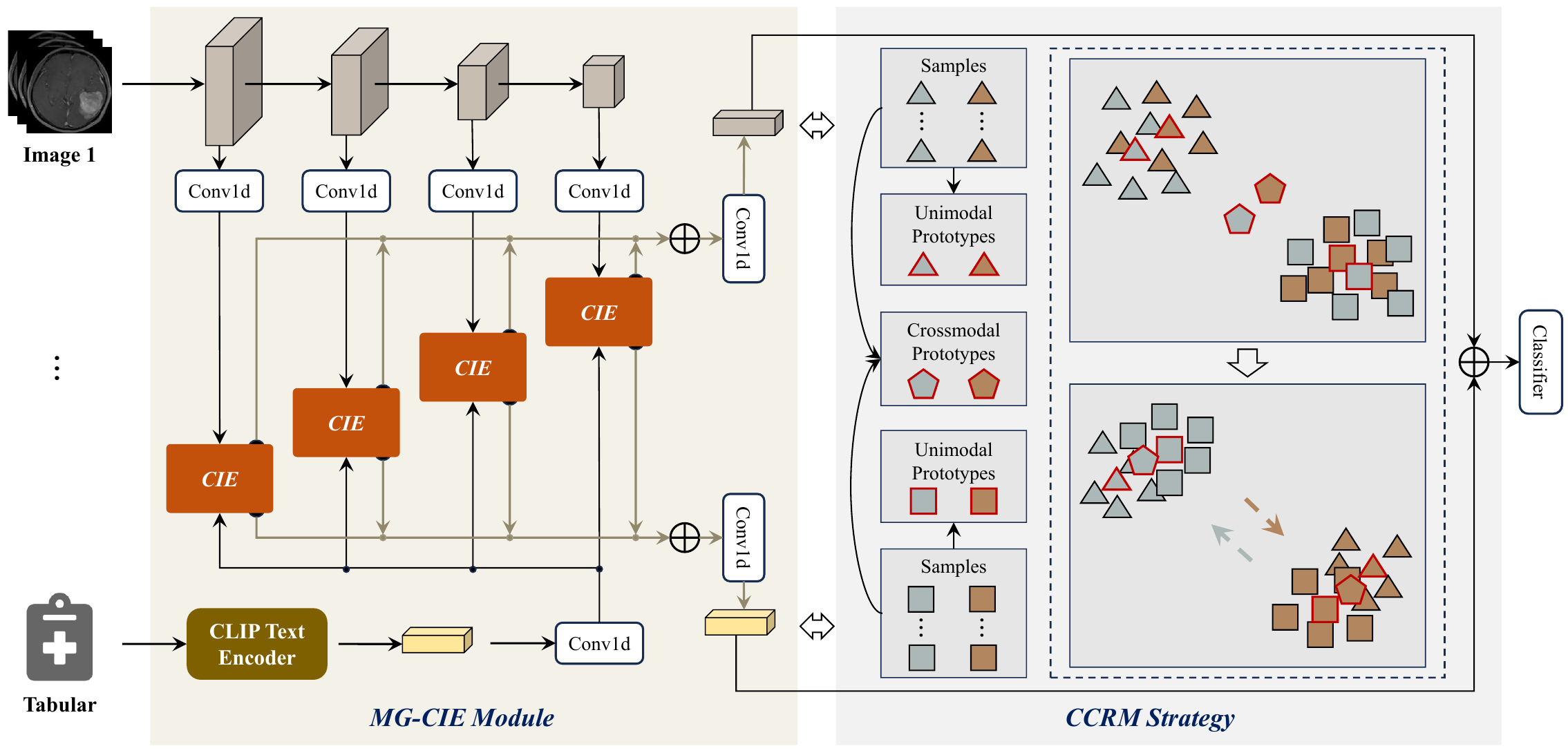}
    \caption{Overview of proposed CFCML framework. For clarity, the figure illustrates a scenario integrating a single image with tabular data. Initially, we extract multi-granularity features from various stages of the image encoder and obtain features from the tabular data using the pretrained CLIP text encoder. Subsequently, the MG-CIE module explore interactions between these multi-granularity features and tabular features to obtain the enhanced unimodal features while preliminary reducing the modality gap. Finally, the CCRM strategy employs class information as a bridge to further narrow the modality gap and extract discriminative features from the crossmodal features. The resulting class-aware crossmodal features are then concatenated to facilitate the final prediction. Specifically, in the CCRM Strategy section, different colors represent samples from distinct classes, while different shapes denote samples from various modalities. The symbol $\oplus$ indicates the concatenation operation.}
    \label{overview}
\end{figure*}

In this paper, we propose a new coarse-to-fine crossmodal learning (CFCML) framework designed to progressively reduce the modality gap by thoroughly exploring intermodal relationships.
The multi-granularity features contain comprehensive information of images, and exploring the relationship between these multi-granularity features and tabular information can yield more complementary information while better mitigating the modality gap.
Therefore, we propose a multi-granularity crossmodal information enhancement (MG-CIE) module. Specifically, we consider the notable differences in the number of tokens between images and tabular data, thereby mapping the token number of the two kinds of modalities to a comparable scale.
We then investigate inter-modal relationships at each level of granularity and ultimately obtain multi-granularity completed unimodal features. For each modality, we integrate the multi-granularity features, ultimately achieving a more comprehensive and robust representation.
Additionally, we propose a novel class-aware relationship mining (CCRM) strategy to further narrow the modality gap using class information and enhance the discriminability of crossmodal features. Specifically, we first establish unimodal prototypes and crossmodal prototypes based on class information.
Subsequently, we develop hierarchical anchor-based contrastive learning strategies that incorporate sample anchor, unimodal prototype anchor and crossmodal prototype anchor.
The objective of these strategies is to cluster samples with the same disease while distancing those with different diseases, thus dissolving the boundaries across modalities while effectively extracting class-aware information.
In summary, our main contributions are summarized as follows:
\begin{itemize}
	\item We develop a coarse-to-fine crossmodal learning framework that progressively reduces the significant modality gap between multimodal images and tabular data by thoroughly exploring the inter-modal relationships.  
	\item We propose a novel MG-CIE module that examines intermodal relationships across multi-granularity features to achieve a more comprehensive unimodal representation while preliminary reduce the modality gap.
	\item We introduce a new CCRM strategy that establishes hierarchical anchor-based contrastive learning strategies, utilizing class information as a bridge to further reduce the modality gap and explore discriminative crossmodal information.
        \item Experiments conducted on both adopted datasets containing multimodal images and tabular data demonstrate the effectiveness and superiority of our method in comparison to state-of-the-art (SOTA) methods.
\end{itemize}

\section{Related Work}
\subsection{Crossmodal learning}
In recent years, numerous studies have focused on CML~\citep{wang2016comprehensive} to achieve more comprehensive and discriminative feature representations. These methods can primarily be classified into three categories: uncertainty-based, feature disentanglement (FD)-based and attention-based.

Several studies have concentrated on uncertainty-based CML for more robust integration of cross modalities~\citep{han2022multimodal,han2022trusted}.
Han et al.
~\citep{han2022trusted} extended the application of evidential deep learning to multi-view learning assigning lower weights to views characterized by high uncertainty during the fusion process.
This type of method lacks interaction between modalities and does not account for the heterogeneity among cross modalities.
Liu et al.~\citep{liu2024dynamic} identified these limitations and proposed the Consistent and Complementary-aware trusted Multi-view Learning (CCML) method, which decoupled the evidence of each view into shared and specific evidence. 

Additionally, numerous studies have focused on the FD-based method~\citep{hu2020disentangled,zheng2022multi,li2023decoupled}. These methods typically decouple each modal feature into a modality-shared and a modality-specific feature. Modality-shared features can reduce the modality gap, while modality-specific features serve as supplementary inputs for information completion.
Zheng et al.~\citep{zheng2022multi} captured the modality-shared and modality-specific features and explored adaptive graph learning to construct patient relationships based on the decoupled features. Li et al.~\citep{li2023decoupled} proposed decoupled multimodal distillation (DMD), which decoupled the information of each modality into modality-irrelevant and modality-exclusive spaces, followed by the application of a graph distillation unit in each space to dynamically enhance the representations of each modality. 

Furthermore, several studies have investigated attention-based CML to minimize the modality gap and complete representation by examining intermodal relationships~\citep{song2021cross,zhu2022multimodal,qiu20243d,zhuang2024glomo,qin2026multimodal}. Song et al.~\citep{song2021cross} developed the crossmodal Attention Blocks to enhance representation of each modality by establishing spatial correspondences between cross modalities. Zhu et al.~\citep{zhu2022multimodal} proposed a triplet attention network to explore complementary information from cross modalities. Qiu et al.~\citep{qiu20243d} learned the global relationships and local associations among cross modalities using attention mechanism. 
Unfortunately, these CML methods primarily focus on high-level information, which is insufficient to fully explore the intermodal relationships, and rarely account for the extraction of task-related information.

Yan et al.~\citep{yan2024causality} proposed the Modality-Relevant Interactive Mining (MRIM) module, which 
learn single-modality and hybrid-modality proxies by incorporating class information and investigate sample-to-sample and sample-to-proxy correlations to capture interactions between modalities and classes. The comparative results presented in Fig. \ref{cr_men} and Fig. \ref{cr_derm7pt} indicate that MRIM achieves only suboptimal performance, suggesting that it faces challenges in adapting to scenarios involving multimodal images and tabular data. Our analysis attributes this limitation to the greater modality differences between image and tabular data compared to those between two image modalities. This disparity complicates the ability of MRIM's contrastive learning algorithm, which relies solely on samples as anchors, to mine class-aware relationships across modalities. Consequently, we propose the establishment of hierarchical anchor-based contrastive learning strategies algorithms which incorporate sample anchor, unimodal prototype anchorand crossmodal prototype anchor to investigate these relationships from various perspectives. 

\subsection{Image-Tabular learning}
In the medical field, the image and tabular data provide complementary information for diagnosing diseases. In recent years, a growing number of studies have focused on the fusion of these two modalities~\citep{cui2023deep}. Several studies~\citep{xia2019multi,el2020multimodal, holste2021end} have employed concatenation operations to fuse image and tabular features, resulting in insufficient interaction between these modalities, which adversely affects the performance of the fused features. 
Duanmu et al.~\citep{duanmu2020prediction} integrated multiple intermediate results from image and tabular data using a channel-wise multiplication operation. Wolf et al.~\citep{wolf2022daft} proposed Dynamic Affine Feature Map Transform (DAFT), which learned an affine transformation for feature maps based on the interaction of image and tabular data. Duenias et al.~\citep{duenias2025hyperfusion} proposed a hypernetwork-based framework for image-tabular fusion that utilized tabular data as priors to improve image network predictions. However, these methods do not account for the heterogeneity between the image and the tabular data. Some studies utilize cross-attention mechanisms to explore the relationships between the two modalities, thereby alleviating the modality gap~\citep{xu2022remixformer,grzeszczyk2023tabattention,guo2024pe,xiong2024multi,ding2024multimodal,jia2024multi,hu2025itcfn}. Wang et al.~\citep{wang2023shared} captured the modality-shared and modality-specific features from image and tabular information using learnable bottleneck tokens. Hager et al.~\citep{hager2023best} aimed to alleviate modality gap through contrastive learning. Xiong et al.~\citep{xiong2024mome} introduced the Mixture of Multimodal Experts (MoME) method, designed to capture intricate intra- and inter-modal interactions between image and tabular data, while incorporating bottleneck features that facilitate bridging gaps between modalities.

However, these studies often do not explore the fusion of multiple images and tabular data and lack mechanisms for filtering out class-irrelevant features during the fusion process.

\section{Method}
\label{section_method}
Let us denote a crossmodal input as $\{X^i,T^i,y^i\}_{i=1}^N$, where $N$ represents the number of samples, $X^i$ denotes the $i$-th multimodal image input, comprising $m$ modalities denoted as $x^i_j (j\in\{1,...,m\})$; $T^i$ represents the $i$-th tabular clinical input containing $t$ attributes; $M=m+1$ is the total number of modalities; $y^i$ is the classification label for the $i$-th sample.

\subsection{Overview}
The overview of the proposed method is illustrated in Fig.~\ref{overview}.
Initially, we extract multi-granularity features from each image based on the multi-stage outputs of the image encoder, and obtain feature embeddings from the tabular data with the pretrained CLIP text encoder. Subsequently, we investigate the correlations between the multi-granularity image features and tabular embeddings by our proposed MG-CIE module, thereby obtaining supplementary information for each modality.
At each granularity, a CIE module generates supplementary information for each modality by exploring the relationships between tabular embeddings and corresponding granularity multimodal image features. This supplementary information is then fused with original features to generate the enhanced features. 
For each modality, we further fuse the multi-granularity enhanced features to derive the final enhanced features. 
Finally, we conduct unimodal and crossmodal prototypes based on class information and employ the CCRM strategy 
to extract class-aware features by exploring relationships among samples and these prototypes.
The objective of the CCRM strategy is to bring samples of the same class closer while pushing samples of different classes apart, thereby transcending the boundaries across modalities. 
Ultimately, the resulting class-aware features are integrated to produce the final prediction.

\subsection{Feature Extraction}
\subsubsection{Multi-granularity feature extraction from images}
Given an input $x$ with dimensions of $\mathbb{R}^{H\times W\times D}$ for 3D image and $\mathbb{R}^{H\times W}$ for 2D image, where $H$, $W$, $D$ denote height, width and depth, respectively. 
We obtain four multi-granularity features $f^s$ (where $s \in\{1,2,3,4\}$) from the outputs of the four stages of the image encoder.
For 3D images, we adopt the nnMamba~\citep{gong2025nnmamba} as encoder and obtain 
$f^s \in\mathbb{R}^{c\times h\times w \times d}$, 
where $c = C*s$ and $C$ denotes the channel dimension of the first stage output, $h = \frac{H}{2^s}$, $w = \frac{W}{2^s}$, $d = \frac{D}{2^s}$. For 2D images, we apply the pretrained Swin Transformer (Swin\_T)~\citep{liu2021swin} as the encoder, following~\citep{xu2022remixformer}, and obtain $f^s \in\mathbb{R}^{c\times h\times w}$, where  $h = \frac{H}{2^{s+1}}$, $w = \frac{W}{2^{s+1}}$. The value of $C$ is set to 16 in nnMamba while 96 in Swin\_T. In the following, 
we will use 3D multimodal images to illustrate the proposed method.

\subsubsection{Tabular embedding}
Previous studies generally input standardized numerical attributes (e.g., age) or convert categorical clinical attributes into one-hot encoding (e.g., sex) to obtain clinical features, 
often overlooking their significance~\citep{cui2023deep}. Inspired by~\citep{xiong2024multi}, we utilized the pretrained CLIP text encoder (ViT-B/32)~\citep{radford2021learning}, which excels in textual feature comprehension, to extract features from tabular data. We have designed several templates to transfer the tabular data into sentences as the input for the text encoder. The details of the templates for different attributes are presented in Table~\ref{template}. Each sentence generated from an attribute will be extracted as a feature with dimensionality of $\mathbb{R}^{1\times 512}$. As a result, we can obtain the extracted tabular features $o\in \mathbb{R}^{t\times 512}$. 
During the training process, the text encoder is frozen.

\begin{figure}
    \centering
    \includegraphics[width=0.7\linewidth]{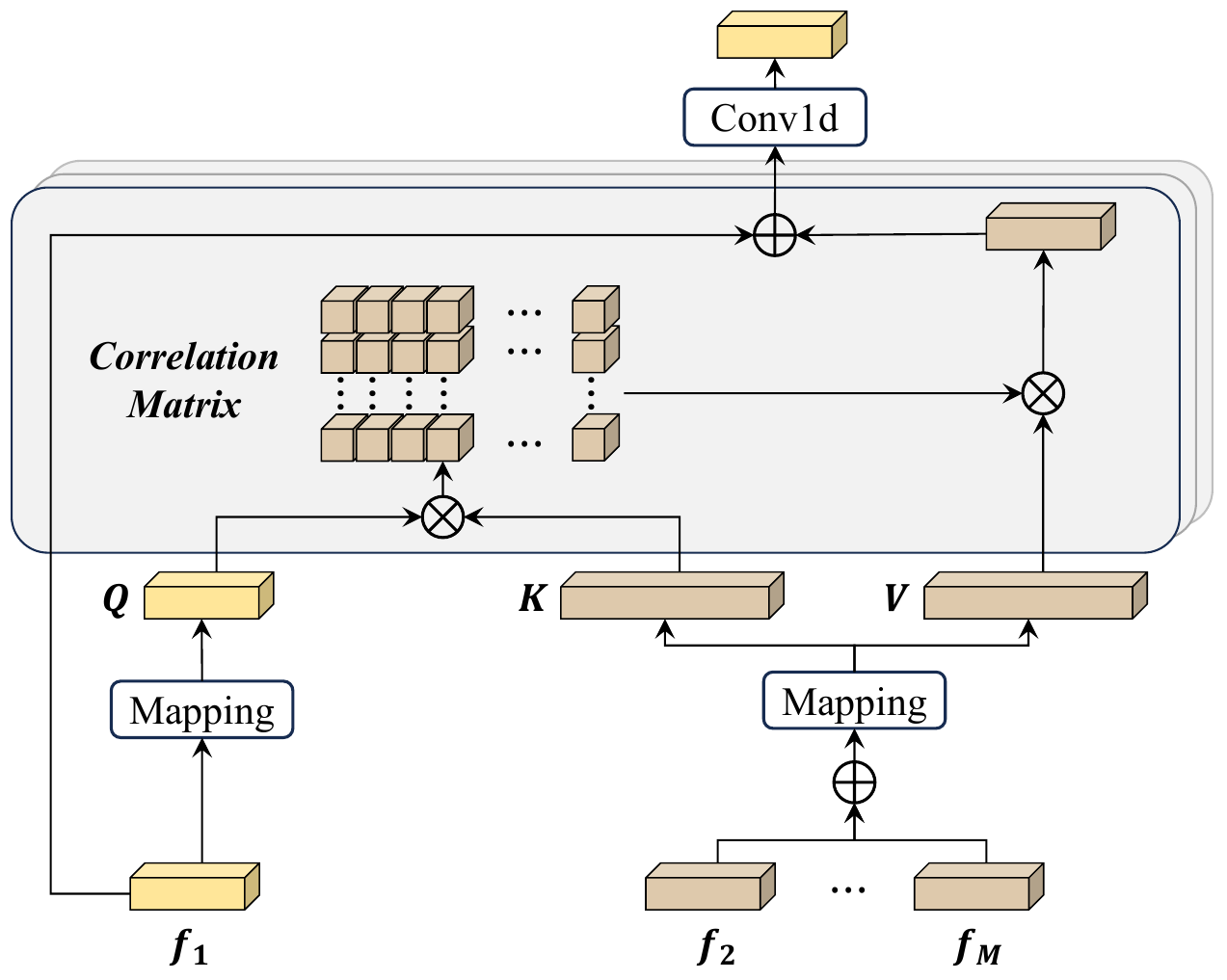}
    \caption{Architecture of the proposed CIE Module.}
    \label{CIE}
\end{figure}
\subsection{Multi-granularity crossmodal information enhancement module}
The deep relationship exploration between tabular embeddings and multi-granularity features from multimodal images is beneficial for reducing modality gap and obtaining more comprehensive unimodal representation.
In fact, there exists a significant disparity in the token numbers between the image and tabular features (e.g., 49,152 tokens for each image at the first granularity compared to 5 tokens for the tabular data in the MEN dataset), resulting in information overload for the images. Consequently, for images and tabular features, we employ distinct adapters to map their tokens to respective predefined numbers, denoted as $n_x$, $n_t$. 
By compressing the image tokens, redundant information can be significantly filtered out, thereby enhancing feature representation.
The adjusted features of each image and tabular data can be obtained as follows.
\begin{equation}
    \hat{f}^s = \Phi^s_f(fc(Re(f^s))),\quad \hat{o} = \Phi_o(fc(o)),
    \label{Eq1}
\end{equation}
where $Re$ represents the operation which converts each granularity feature $f^s$ into a token sequence with dimensions of $\mathbb{R}^{hwd\times c}$; $fc(\cdot)$ denotes a fully connected layer that maps the multi-granularity image features or tabular features to a unified dimension $C_d=128$; $\Phi(\cdot)$ represents the $Conv1d$ operation that maps the number of tokens to a predefined number. Specifically, the input channel of $Conv1d$ corresponds to the original token count for image or tabular features, while the output channel corresponds to the predefined numbers.

We subsequently investigate the correlation between adjusted tabular features and multimodal image features across different granularity using MG-CIE module, as shown in Fig.~\ref{overview}. As an illustrative example, we demonstrate the correlation exploration process between the tabular features $\hat{t}$ and the first granularity multimodal image features $\{\hat{f}_1^1, .., \hat{f}_m^1\}$. As depicted in Fig.~\ref{CIE}, 
we sequentially designate one feature as the primary feature $f_b$ and treat the concatenated tokens from all other features as the auxiliary feature $f_a$. Then, we perform multi-head cross-attention between the primary and auxiliary feature to capture the additional information for the primary feature. In this context, the primary feature serves as the query, while the auxiliary information acts as both the key and value. Mathematically, this process can be formulated as:
\begin{align}
    &Q = f_bW_Q,\,K = f_aW_K,\,V = f_aW_V,\\
    &M_{corr} = softmax(\frac{QK^T}{\sqrt{C_d/N_h}}),\\
    &f_b^{sup} = M_{corr}V,\quad \tilde{f_b} = \Phi_p(Cat(f_b,f_b^{sup})),
    \label{Eq4}
\end{align}
where $W_Q, W_K, W_V \in\mathbb{R}^{C_d\times C_d}$ are learnable parameters; $N_h$ denotes the number of heads; $M_{corr}$ represents the correlation matrix between tokens of the primary feature and those of the auxiliary feature; $f_b^{sup}$ denotes the supplementary features for the primary features; $Cat$ signifies the token concatenation operation; $\Phi_p$ represents the $Conv1d$ operation that maps the number of concatenated token to a predefined number; and $\tilde{f_b}$ indicates the enhanced features. As a result, we obtain the enhanced multimodal features $\{\tilde{f}_1^1, .., \tilde{f}_m^1\}$ and the enhanced tabular feature $\tilde{o}^1$ at the first granularity. In this way, we finally obtain the enhanced multimodal image features and enhanced tabular features at different granularity.

We then fuse the enhanced features from different granularities for each modality as follows:
\begin{equation}
    P = \Phi^{mg}(Cat(\tilde{p}^1,\tilde{p}^2,\tilde{p}^3,\tilde{p}^4),\quad p\in\{f,o\},
    \label{Eq5}
\end{equation}
where $\tilde{p}^i$ represents the enhanced feature at the $i$-th granularity, while $P\in\{F,O\}$ denotes the fused features from multiple granularities; $\Phi^{mg}$ indicates the $Conv1d$ operation that maps the number of concatenated token from multi-granularity enhanced features to the predefined number ($N_x$ for image, $N_t$ for tabular data). Ultimately, we obtain final enhanced multimodal image features $\{F_1,...,F_m\}$ and tabular feature $O$. For the convenience of the following representation, we have established an enhanced feature set $\mathcal{Z}=\{F_1,...,F_m,O\}$.

\begin{figure}
    \centering
    \includegraphics[width=\linewidth]{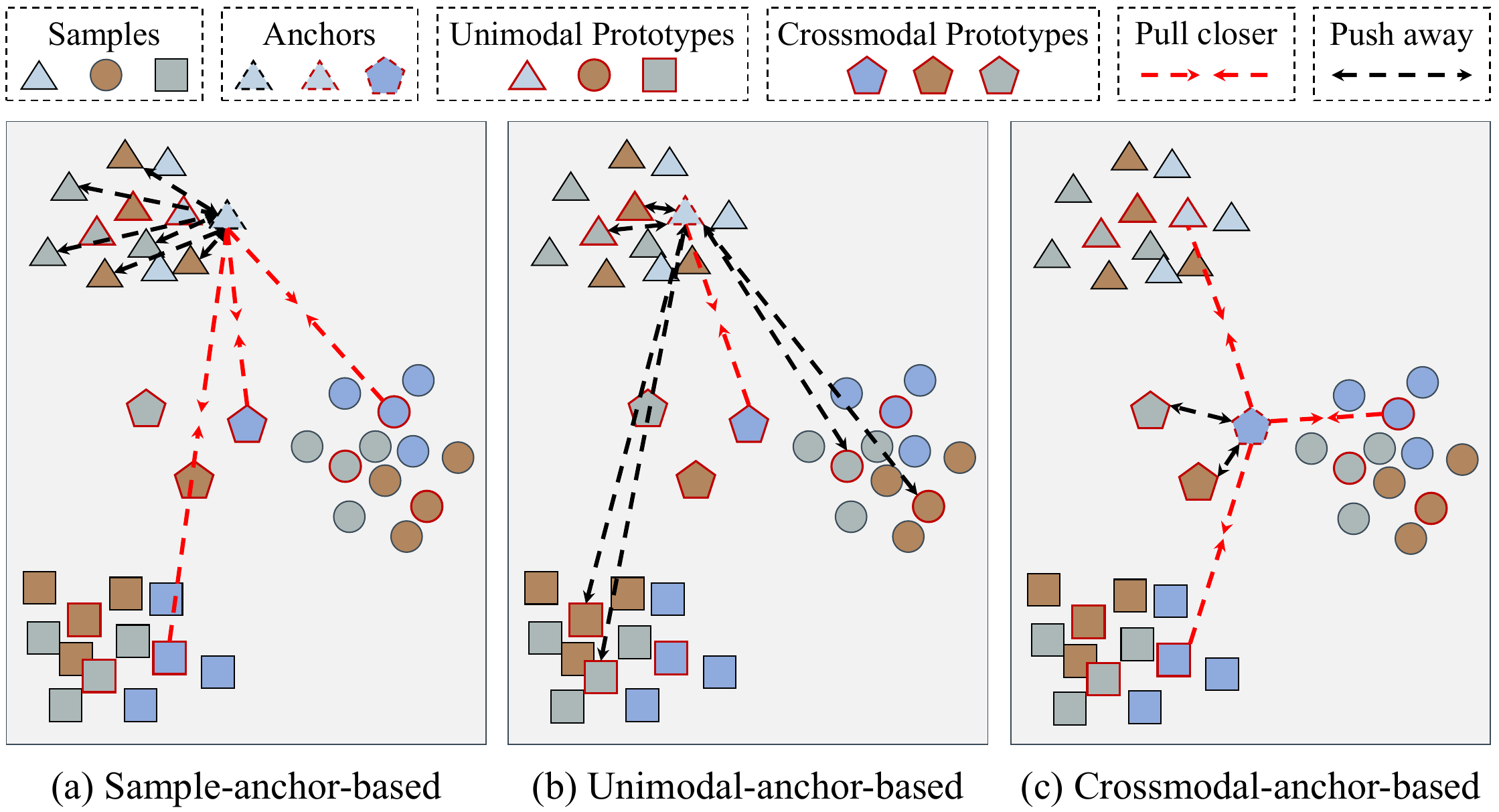}
    \caption{Illustration of the CCRM strategy across three modalities and three class conditions.
    CCRM encompasses sample-anchor-based, unimodal-anchor-based, and crossmodal-anchor-based CL strategies. Specifically, in each strategy, we select an anchor sample and identify its positive samples (connected by the red dashed line) and negative samples (connected by the black dashed line). The objective is to minimize the distance between the anchor and positive samples while maximizing the distance between the anchor and negative samples.}
    \label{CCRM}
\end{figure}
\subsection{Class-aware crossmodal relationship mining strategy}
The MG-CIE module can preliminarily address the modality gap, and each enhanced unimodal representation contains numerous class-irrelevant features, which may adversely affect overall classification performance.
To address these limitations, we propose a new Class-aware Crossmodal Relationship Mining (CCRM) strategy, which utilizes class information as a bridging mechanism to further mitigate the modality gap while simultaneously extracting discriminative features. 

Fig.~\ref{CCRM} presents examples of three modalities and three classes. As illustrated in this figure, we first establish a crossmodal prototype $cp$ for samples belonging to each class, as well as a unimodal prototype $up$ for samples within each modality that correspond to each class.
\begin{equation} 
cp^{l} = \frac{\sum_{i=1}^{N}\sum_{j=1}^{M} \mathcal{Z}^{i}_{j} \mathbbm{1}[y^{i} = l]}{\sum_{i=1}^{N}\sum_{j=1}^{M} \mathbbm{1}[y^{i} = l]},
\label{Eq6}
\end{equation}
\begin{equation} 
up^{l}_j = \frac{\sum_{i=1}^{N} \mathcal{Z}^{i}_{j} \mathbbm{1}[y^{i} = l]}{\sum_{i=1}^{N} \mathbbm{1}[y^{i}=l]},\ j\in\{1,...,M\},
\label{Eq7}
\end{equation}
where $l\in N_c$ denotes the index of class number, $\mathcal{Z}_j^i$ represents the enhanced feature of the $j$-th modality for the $i$-th sample, and $\mathbbm{1}[\cdot]$ is the indicator function. 

Due to inherent characteristics, samples from the same modality tend to cluster together, while those from different modalities are distributed further apart, resulting in significant modality gap while neglecting class-aware information.
Therefore, we introduce hierarchical anchor-based contrastive learning (CL) strategies to further reduce the modality gap and mine class-aware crossmodal features, which encompasses three components: sample-anchor-based, unimodal proto-anchor-based and crossmodal proto-anchor-based.

\subsubsection{Sample-anchor-based CL strategy}
To differentiate samples of various classes within the same modality and enhance the class-aware information associated with each sample, we establish a sample-anchor-based CL strategy. Specifically, each sample is treated as an anchor, with unimodal prototypes and crossmodal prototype sharing the same class forming the positive sample set, denoted as $S_{sam^+}$, while samples with different labels from the same modality comprise the negative sample set, denoted as $S_{sam^-}$. This strategy is mathematically expressed as:
\begin{equation}  
    \mathcal{L}_{sam} = -log \sum_{i=1}^{N} \sum_{j=1}^{M} \frac{Op_+(\mathcal{Z}^i_j, S_{sam^+})}{Op_+(\mathcal{Z}^i_j, S_{sam^+}) + Op_-(\mathcal{Z}^i_j, S_{sam^-})},
    \label{Eq8}
\end{equation}
where $Op_+(ar,S_+)$ represents the cumulative similarity between the anchor $ar$ and each sample in the positive sample set $S_+$; $Op_-(ar,S_-)$ denotes the cumulative similarity between the anchor $ar$ and each sample in the negative sample set $S_-$. These operations can be expressed as:
\begin{equation}
    Op_*(ar, S_+)=\sum_{k=1}^{N_*}\exp(CS(ar, S^k_*)) / \tau,
\end{equation}
where $*\in\{+,-\}$, $N_+$ and $N_-$ denote the number of samples in the positive and negative sample sets, respectively; $CS(\cdot, \cdot)$ signifies the cosine similarity function, $\tau$ represents the temperature parameter.

\subsubsection{Unimodal proto-anchor-based CL strategy}
We further examine the relationships between unimodal prototypes to capture discriminative information. Specifically, we designate each unimodal prototype as the anchor, with crossmodal prototype that belongs to the same class forming the positive sample set $S_{up^+}$, while unimodal prototypes from different classes serve as the negative sample set $S_{up^-}$. The mathematical representation of this strategy is as:
\begin{equation}  
    \mathcal{L}_{up} = -log \sum_{l=1}^{N_c} \sum_{j=1}^{M} \frac{Op_+(up_j^l, S_{up^+})}{Op_+(up_j^l, S_{up^+}) + Op_-(up_j^l, S_{up^-})}.
    \label{Eq9}
\end{equation}

\subsubsection{Crossmodal proto-anchor-based CL strategy}
To further enhance the separation between different classes and improve the discriminative power of crossmodal features, we propose a crossmodal proto-anchor-based CL strategy. Specifically, each crossmodal prototype serves as an anchor, with all unimodal prototypes sharing the same label comprising the positive sample set $S_{cp^+}$, while other crossmodal prototypes constitute the negative sample set $S_{cp^-}$. This strategy can be mathematically formulated as:
\begin{equation}  
    \mathcal{L}_{cp} = -log \sum_{l=1}^{N_c} \sum_{j=1}^{M} \frac{Op_+(cp_j^l, S_{cp^+})}{Op_+(cp_j^l, S_{cp^+}) + Op_-(cp_j^l, S_{cp^-})}.
    \label{Eq10}
\end{equation}

Finally, class-aware information is filtered from each enhanced feature through backpropagation, enabling the acquisition of final fused features by concatenating all extracted class-aware features. To obtain the final prediction $y'$, we utilize a Multi-Layer Perceptron (MLP) as the classifier. The cross-entropy (CE) loss acts as the supervisory signal for prediction. The classification loss $\mathcal{L}_{cls}$ is defined as follows:
\begin{equation}
\mathcal{L}_{cls} = CE(y',y).
\label{Eq12}
\end{equation}
The overall loss $\mathcal{L}$ is characterized as the weighted sum of the previously mentioned losses,
\begin{equation}
\mathcal{L} = \mathcal{L}_{cls} + \alpha\mathcal{L}_{sam}+\beta \mathcal{L}_{up}+\gamma\mathcal{L}_{cp},
\label{Eq13}
\end{equation}
where $\alpha$, $\beta$ and $\gamma$ are employed as balance factors.

The description of the algorithm for our proposed CFCML framework is summarized in
\textbf{Algorithm~\ref{algorithm1}}.

\begin{algorithm}[t]
    \caption{CFCML Algorithm}
    \label{algorithm1}
    \KwIn{Multimodal images $X^i$ and tabular data $T^i$; Target class $y^i$.}
    
    \For{e = 1; e $\leq$ Epoch; e + + }{
        1. Extract multi-granularity features $f^s_j$ from multimodal images and $o^s$ from tabular data, where $j\in\{1,...,m\}, s\in\{1,2,3,4\}$; \\
        2. Obatain the adjusted multi-granularity multimodal image features $\hat{f}^s_j$ and tabular features $\hat{o}^s$ as Eq.\ref{Eq1}; \\
        3. Get enhanced multimodal features $\tilde{f}_1^s, .., \tilde{f}_m^s$ and the enhanced tabular feature $\tilde{o}^s$ at multiple levels of granularity as Eq.\ref{Eq4}; \\
        4. Obtain the final enhanced multimodal image features $F_1, ..., F_m$ and tabular feature $O$ from multi-granularity features as Eq.\ref{Eq5}; \\
        5. Calculate the crossmodal prototype $cp$ and the unimodal prototype $up$ as Eq.\ref{Eq6} and Eq.\ref{Eq7}; \\
        6. Conduct relationship mining including 
        sample-anchor-based, unimodal proto-anchor-based and crossmodal proto-anchor-based approaches as Eq.\ref{Eq8}-\ref{Eq10}; \\
        7. Concatenate the extracted class-aware features and obtain the final prediction $y'$; \\
        8. Compute the classification loss and overall loss as Eq.\ref{Eq12} and Eq.\ref{Eq13}; \\
    }  
\end{algorithm}
\section{Experiments}
\subsection{Datasets}
\begin{figure}
    \centering    
    \includegraphics[width=\linewidth]{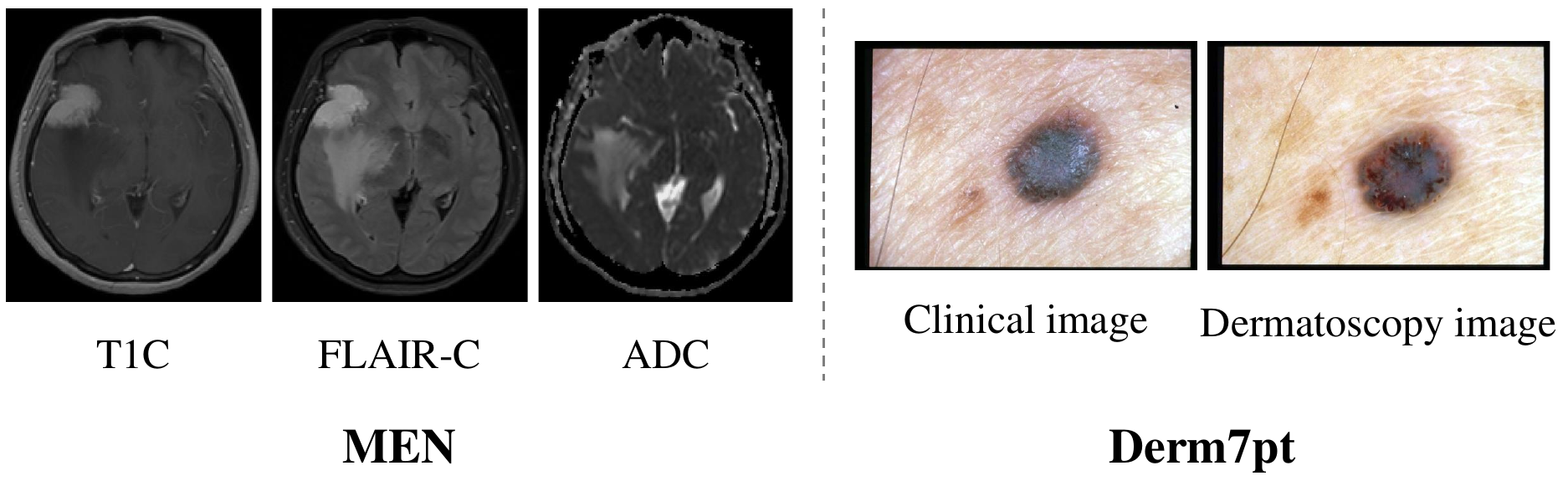}
    \caption{Display of multimodal images from the adopted datasets.}
    \label{data_show}
\end{figure}

To validate the effectiveness of our proposed method, we conducted experiments using two crossmodal datasets, including a private dataset and a public dataset. Sample instances from these datasets are presented in Fig.~\ref{data_show}.
\subsubsection{MEN dataset}
The MEN dataset was collected from the Brain Medical Center of Tianjin University, Tianjin Huanhu Hospital\footnote{The Ethical Committee of Tianjin Huanhu Hospital has granted approval for the scientific research involving multimodal MRIs and clinical data and has waived the necessity for informed patient consent (Jinhuan Ethical Review No. 2022-046)}.
In total, we gathered multimodal magnetic resonance imaging (MRI) and clinical information from 796 patients diagnosed with meningiomas of three grades: Grade 1 (G1), Grade 2 with invasion (G2inv), and Grade 2 without invasion (G2ninv). The dataset comprises 650 patients with G1, 60 patients with G2inv, and 86 patients with G2ninv. We adopted three MRI series for each patient: Contrast-Enhanced T1 series (T1C), Contrast-Enhanced T2 FLAIR series (FLAIR-C), and Apparent Diffusion Coefficient series (ADC). The clinical information consists of six attributes: sex, age, tumor area, edema area, tumor location, and the value of the apparent diffusion coefficient. Following~\citep{liu2025completed}, we zero-padded the regions of interest (ROIs) in MRIs, including the tumor and edema area, to squares and resized them to dimensions of 24*128*128 for model input. 

\subsubsection{Derm7pt dataset}
The Derm7pt~\citep{kawahara2018seven} is a publicly available crossmodal skin lesion analysis dataset.
Following~\citep{patricio2023coherent,hou2024concept}, we focused on predicting disease categories, specifically melanoma (MEL) and nevus (NEV), and filtered the original dataset to obtain a subset of 827 cases, each containing multimodal images and clinical data. There are 346 cases in the training set (90 NEV and 256 MEL cases), 161 cases in the validation set (61 NEV and 100 MEL cases), and 320 cases in the test set (101 NEV and 219 MEL cases), respectively. For each case, we adopted two multimodal images, consisting of clinical and dermatoscopy images, alongside five attributes, including sex, management, lesion location, lesion elevation, and level of diagnostic difficulty. All images were resized to dimensions of 224*224*3 for model input.

\subsection{Evaluation metrics}
Following~\citep{liu2025completed}, we adopted 7 metrics to validate the effectiveness of the proposed method for the MEN dataset, including Accuracy (ACC), Accuracy of G1 (ACC\_G1), Accuracy of G2inv (ACC\_G2inv), Accuracy of G2ninv (ACC\_G2ninv), weighted F1 score (weighted-F1), macro F1 score (macro-F1) and Area Under the Curve (AUC). For the Derm7pt dataset, we employed 7 metrics as well, which included Sensitivity (SEN), Specificity (SPE), Accuracy (ACC), G-mean, Balanced Accuracy (Ba\_ACC)~\citep{brodersen2010balanced}, Area Under the Precision-Recall Curve (AUPRC), and AUC.
In the statistical analysis, we utilized the Wilcoxon signed-rank test~\citep{wilcoxon1992individual} to compare the metrics of our proposed method with SOTA methods.

\begin{table}[t]  
\centering  
\caption{The templates for different attributes. In each template, the first \{\} represents the attribute, and the second \{\} represents the attribute value.}  
\label{template}  
\resizebox{0.45\textwidth}{!}{  
\begin{tabular}{cc}  
\hline
\multirow{2}{*}{\textbf{Attributes}}    & \multirow{2}{*}{\textbf{Templates}}             \\
                                        &                                                 \\ \hline
management                              & The \{\} for the patient is \{\}                \\ \hdashline
sex                                     & \multirow{2}{*}{The \{\} of patient is \{\}}    \\
age                                     &                                                 \\ \hdashline
tumor area                              & \multirow{3}{*}{The \{\} in the brain is \{\}.} \\
edema area                              &                                                 \\
tumor location                          &                                                 \\ \hdashline
lesion location                         & \multirow{4}{*}{The \{\} is \{\}}               \\
lesion elevation                        &                                                 \\
level of diagnostic difficulty          &                                                 \\
value of apparent diffusion coefficient &                                                 \\ \hline
\end{tabular}}
\end{table}
\subsection{Experiment setup and implementation details}
For both adopted datasets, we utilized the Adam optimizer with a weight decay of $1e-4$ to update the model. The training was conducted over $50$ epochs. We implemented a linear warm-up strategy during the first $5$ epochs, gradually increasing the learning rate from zero to its initial value. Additionally, we applied a learning rate decay strategy that reduced the learning rate to 80\% of its previous value every $5$ epochs. The dropout rate was set to $0.5$.  The temperature parameter $\tau$ is set as $0.07$~\citep{wu2018unsupervised, he2020momentum}. We transferred tabular attributes to sentences using the templates presented in Table~\ref{template}.
For the MEN dataset, we employed 3-fold cross-validation to evaluate performance. The learning rate was set to $5e-4$, and the batch size was established at $36$. Data augmentation techniques including Gaussian noise, random crop, random flip, and random erasing~\citep{zhong2020random}, were applied to the training images. In the case of the Derm7pt dataset, we trained the model three times using different seeds, utilizing training, validation, and testing data pre-divided by the creator. We initialized the learning rate value at $1e-4$ and set the batch size to 64. Following~\citep{tang2022fusionm4net}, we employed random vertical and horizontal flips, shifts, and distortions as data augmentation operations.
The predefined token numbers for image and tabular data, \{$n_x$, $n_t$\} were set to \{48, 16\} for the MEN dataset and \{32, 24\} for the Derm7pt dataset. The ablation studies regarding the token number settings are discussed in the Section~\ref{discussion}.
We adjusted the balance factors to ensure that the initial magnitudes of the other loss components (i.e. $\mathcal{L}_{sam}, \mathcal{L}_{up}, \mathcal{L}_{cp}$) were comparable to that of the task loss ($\mathcal{L}_{cls}$). Specifically, the factors $\alpha$, $\beta$, $\gamma$ were set to $0.06$, $0.04$ and $0.24$ for the MEN dataset, and to $0.04$, $0.06$, $0.18$ for the Derm7pt dataset, respectively. All experiments were conducted using Python 3.10 with the PyTorch toolkit 2.1 on a platform equipped with an NVIDIA GeForce RTX 3090 GPU.

\begin{table*}[t]
\centering
\caption{The comparison results on the MEN dataset (mean$\pm$standard deviation). The best results for each metric are highlighted in \textbf{bold}.  * indicates that our proposed method achieves statistically significant improvements over the other compared methods (p-value<0.05). The complexity, including the number of parameters (M) and GFLOPs (G), is displayed on the right. The term MRIM+ denotes the integration of the proposed MG-CIE module into the comparative method MRIM, while Proposed- indicates the replacement of the MG-CIE module with the single-granularity CIE at the final stage of the encoder (SG-CIE).}
\label{cr_men}
\renewcommand{\arraystretch}{1.3}
\resizebox{\textwidth}{!}{
\begin{tabular}{c|lllllll|cc}
\hline
\multirow{2}{*}{Methods} & \multicolumn{7}{c|}{Metrics}                                                                                   & \multicolumn{2}{c}{Complexity} \\
                         & \multicolumn{1}{c}{ACC} & \multicolumn{1}{c}{ACC\_G1} & \multicolumn{1}{c}{ACC\_G2inv} & \multicolumn{1}{c}{ACC\_G2ninv} & \multicolumn{1}{c}{weighted-F1} & \multicolumn{1}{c}{macro-F1} & \multicolumn{1}{c|}{AUC} & Param     & GFLOPs     \\ \hline
ETMC~\cite{han2022trusted}                     & 88.66$ \pm $4.74*            & 93.99$ \pm $7.22*                & 66.94$ \pm $8.68*                   & 63.73$ \pm $15.01*                   & 89.02$ \pm $3.40*                    & 75.39$ \pm $3.54*                 & 90.19$ \pm $0.98*             & 10.50         & 18.51          \\
DMD~\cite{li2023decoupled}                      & 92.21$ \pm $0.51*            & 96.46$ \pm $1.42*                & 91.49$ \pm $7.70                    & 60.72$ \pm $9.45*                    & 92.13$ \pm $0.25*                    & 81.34$ \pm $1.15*                 & 96.25$ \pm $0.73*             & 171.66        & 18.74          \\
MVCNet~\cite{guo2024pe}                & 93.84$ \pm $1.33*            & \textbf{97.53$ \pm $2.72}                 & 86.72$ \pm $7.35*                   & 71.11$ \pm $6.74*                    & 93.81$ \pm $1.12*                    & 85.84$ \pm $2.50*                 & 96.83$ \pm $1.05*             & 10.67         & 18.51          \\
CCML~\cite{liu2024dynamic}                     & 91.33$ \pm $1.06*            & 96.92$ \pm $1.63                 & 81.22$ \pm $13.45*                  & 56.19$ \pm $14.02*                   & 90.91$ \pm $0.92*                    & 79.50$ \pm $3.00*                  & 95.41$ \pm $1.36*             & 10.36         & 18.51          \\
GLoMo~\cite{zhuang2024glomo}                    & 94.35$ \pm $0.95*            & 96.62$ \pm $0.69                 & 86.56$ \pm $3.41*                   & 82.62$ \pm $5.82*                    & 94.50$ \pm $0.91*                     & 87.40$ \pm $2.70*                  & 97.04$ \pm $1.19*             & 12.85         & 18.60          \\
MRIM~\cite{yan2024causality}                     & 90.32$ \pm $0.93*            & 95.07$ \pm $2.64*                & 76.12$ \pm $16.64*                  & 64.52$ \pm $21.31*                   & 90.29$ \pm $1.04*                    & 78.78$ \pm $1.24*                 & 95.00$ \pm $0.40*              & 10.39         & 18.51          \\
MRIM+                    & 92.57$ \pm $2.63*            & 94.92$ \pm $2.83*                & 88.05$ \pm $13.33*                  & 77.86$ \pm $12.64*                   & 93.02$ \pm $2.32*                    & 82.88$ \pm $4.87*                 & 96.21$ \pm $0.56*             & 35.98         & 19.07          \\ \hline
\textbf{Proposed-}                & 92.97$ \pm $0.51             & 95.08$ \pm $2.69                 & 91.57$ \pm $3.19                    & 77.70$ \pm $14.72                     & 93.24$ \pm $0.17                     & 84.99$ \pm $3.33                  & 97.80$ \pm $0.33               & 10.71         & 18.07          \\
\textbf{Proposed}                 & \textbf{95.61$ \pm $0.56}             & 96.77$ \pm $1.23                 & \textbf{93.24$ \pm $3.26}                    & \textbf{88.57$ \pm $7.42}                     & \textbf{95.76$ \pm $0.52}                     & \textbf{91.13$ \pm $1.28}                  & \textbf{98.57$ \pm $0.12}              & 35.98         & 19.07          \\ \hline
\end{tabular}}
\end{table*}
\begin{table*}[t]
\centering
\caption{The comparison results on the Derm7pt dataset.}
\label{cr_derm7pt}
\renewcommand{\arraystretch}{1.3}
\resizebox{\textwidth}{!}{
\begin{tabular}{c|lllllll|cc}
\hline
\multirow{2}{*}{Methods} & \multicolumn{7}{c|}{Metrics}                                                                           & \multicolumn{2}{c}{Complexity} \\
                         & \multicolumn{1}{c}{SEN}          & \multicolumn{1}{c}{SPE}          & \multicolumn{1}{c}{ACC}          & \multicolumn{1}{c}{G\_Mean}      & \multicolumn{1}{c}{Ba\_ACC}      & \multicolumn{1}{c}{AUPRC}        & \multicolumn{1}{c|}{AUC}          & Param          & GFLOPs        \\ \hline
ETMC~\cite{han2022trusted}                     & 77.56$ \pm $8.42  & 73.98$ \pm $5.54* & 75.11$ \pm $2.90* & 75.57$ \pm $3.02* & 75.77$ \pm $3.22* & 52.13$ \pm $3.44* & 83.06$ \pm $1.67* & 28.69          & 44.66         \\
DMD~\cite{li2023decoupled}                      & 69.97$ \pm $5.80* & 88.89$ \pm $2.16  & 82.92$ \pm $0.36* & 78.79$ \pm $2.38* & 79.43$ \pm $1.82* & 61.56$ \pm $0.99* & 89.61$ \pm $1.06* & 184.49         & 44.76         \\
MVCNet~\cite{guo2024pe}                & 73.27$ \pm $3.57* & 86.45$ \pm $1.90* & 82.29$ \pm $0.65* & 79.56$ \pm $1.24* & 79.86$ \pm $1.07* & 60.77$ \pm $1.12* & 88.88$ \pm $0.50* & 28.89          & 44.66         \\
CCML~\cite{liu2024dynamic}                     & 69.97$ \pm $2.06* & 83.71$ \pm $2.16* & 79.38$ \pm $1.90* & 76.53$ \pm $1.84* & 76.84$ \pm $1.84* & 56.05$ \pm $2.80* & 86.63$ \pm $2.41* & 28.58          & 44.66         \\
GLoMo~\cite{zhuang2024glomo}                    & \textbf{79.87$ \pm $3.48}  & 78.69$ \pm $8.12* & 79.06$ \pm $4.81* & 79.15$ \pm $3.17* & 79.28$ \pm $3.03* & 57.48$ \pm $5.61* & 88.77$ \pm $1.85* & 30.67          & 44.69         \\
MRIM~\cite{yan2024causality}                     & 71.95$ \pm $5.45* & 86.00$ \pm $4.60* & 81.56$ \pm $1.44* & 78.55$ \pm $0.91* & 78.97$ \pm $0.46* & 59.65$ \pm $1.82* & 88.79$ \pm $1.14* & 26.60           & 44.67         \\
MRIM+              & 75.91$ \pm $5.08  & 85.85$ \pm $5.19* & 82.71$ \pm $2.43* & 80.62$ \pm $1.67* & 80.88$ \pm $1.56* & 61.93$ \pm $3.48* & 89.02$ \pm $1.79* & 30.63          & 79.79         \\ \hline
\textbf{Proposed-}                & 73.60$ \pm $3.02  & 88.58$ \pm $1.99  & 83.85$ \pm $1.41  & 80.72$ \pm $1.65  & 81.09$ \pm $1.54  & 63.47$ \pm $2.54 & 89.70$ \pm $1.62  & 28.81          & 44.68         \\
\textbf{Proposed}                 & 75.91$ \pm $3.48  & \textbf{89.49$ \pm $2.09}  & \textbf{85.21$ \pm $1.41}  & \textbf{82.40$ \pm $1.75}  & \textbf{82.70$ \pm $1.63}   & \textbf{66.07$ \pm $2.64}  & \textbf{90.52$ \pm $0.96} & 30.63          & 79.79         \\ \hline
\end{tabular}}
\end{table*}

\subsection{Quantitative results}
To showcase the superiority of our proposed method, we identified six SOTA methods, including (1) the uncertainty-based methods, i.e., ETMC~\citep{han2022trusted}, CCML~\citep{liu2024dynamic}, (2) the feature disentanglement-based methods like DMD~\citep{li2023decoupled}, (3) the attention-based methods, such as MVCNet\footnote{MVCNet was originally designed to fuse a single image with tabular data. To adapt it to our multimodal images and tabular data fusion scenario, we fuse the extracted features of the multimodal image through a channel concatenation operation and subsequently interact with the tabular data.}~\citep{guo2024pe}, GLoMo~\citep{zhuang2024glomo}, and (4) MRIM~\citep{yan2024causality}. 
To ensure a fair comparison, we configured the encoders for multimodal images and tabular data to be consistent with those of the proposed method. 

The comparison results on the MEN dataset are presented in Table~\ref{cr_men}.
Notably, our proposed method outperforms all comparison methods across nearly all metrics, achieving improvements of $1.26\%$ in ACC,$1.75\%$ in ACC\_G2inv, $5.95\%$ in ACC\_G2ninv, $1.26\%$ in weighted-F1, $3.73\%$ in macro-F1 and $1.53\%$ in AUC compared to the SOTA methods with the best results. In Table~\ref{cr_derm7pt}, we further validate the proposed method on the public Derm7pt dataset, where it also achieves superior results across almost all metrics compared to the SOTA methods. Specifically, our proposed method demonstrates improvements of $0.6\%$ in SPE, $2.29\%$ in ACC, $2.84\%$ in G\_Mean, $2.84\%$ in Ba\_ACC, $4.51\%$ in AUPRC and $0.91\%$ in AUC.

Furthermore, the results of the statistical tests conducted on both datasets, as presented in Table~\ref{cr_men} and Table~\ref{cr_derm7pt}, provide additional evidence that our proposed method significantly surpasses the comparison methods across the majority of metrics. In summary, the high accuracy attained in each category suggests that the proposed method is capable of effectively extracting more discriminative crossmodal features.

\subsection{Ablation studies}
We conduct ablation experiments to justify the effectiveness of the two components in the proposed method, including the MG-CIE module and the CCRM strategy. The ablation results are shown in Table~\ref{ab_men} and Table~\ref{ab_derm7pt} for the MEN and Derm7pt datasets, respectively.

\subsubsection{Effectiveness of the MG-CIE module}
Comparing the first two rows in Table~\ref{ab_men}, it is evident that the MG-CIE module significantly enhances most metrics on the MEN dataset, particularly with a $3.2\%$ improvement in the macro-F1 metric. A similar phenomenon is observed on the derm7pt dataset, as shown in Table~\ref{ab_derm7pt}. These improvements indicate that the MG-CIE module has obtained more comprehensive unimodal features by exploring the relationships among cross modalities.
\begin{table}[t]
\centering
\caption{The ablation study on the MEN dataset.}
\label{ab_men}
\Large
\resizebox{0.48\textwidth}{!}{
\begin{tabular}{ccccccccc}
\hline
\multicolumn{2}{c}{Proposed} & \multirow{2}{*}{ACC} & \multirow{2}{*}{\makecell[c]{\vspace{-1mm} ACC\\ \_G1}} & \multirow{2}{*}{\makecell[c]{\vspace{-1mm} ACC\\ \_G2inv}} & \multirow{2}{*}{\makecell[c]{\vspace{-1mm} ACC\\ \_G2ninv}} & \multirow{2}{*}{\makecell[c]{\vspace{-1mm} weighted\\ -F1}} & \multirow{2}{*}{\makecell[c]{\vspace{-1mm} macro\\ -F1}} & \multirow{2}{*}{AUC} \\ \cline{1-2}
MG-RIE         & CCRM        &                      &                          &                             &                              &                               &                            &                      \\ \hline
$\times$              & $\times$           & 91.32                & 95.84                    & 82.06                       & 64.13                        & 91.37                         & 80.95                      & 95.77                \\
\checkmark              & $\times$           & 92.84                & 96.00                    & 84.73                       & 74.44                        & 92.99                         & 84.15                      & 95.90                \\
$\times$              & \checkmark           & 91.70                & 94.61                    & 86.48                       & 73.49                        & 92.04                         & 82.42                      & 96.02                \\
\checkmark              & \checkmark           & \textbf{95.61}                & \textbf{96.77}                    & \textbf{93.24}                       & \textbf{88.57}                        & \textbf{95.76}                         & \textbf{91.13}                      & \textbf{98.57}                \\ \hline
\end{tabular}}
\end{table}
\begin{table}[t]
\centering
\caption{The ablation study on the Derm7pt dataset.}
\label{ab_derm7pt}
\Large
\resizebox{0.48\textwidth}{!}{
\begin{tabular}{ccccccccc}
\hline
\multicolumn{2}{c}{Proposed} & \multirow{2}{*}{SEN} & \multirow{2}{*}{SPE} & \multirow{2}{*}{ACC} & \multirow{2}{*}{G\_Mean} & \multirow{2}{*}{Ba\_ACC} & \multirow{2}{*}{AUPRC} & \multirow{2}{*}{AUC} \\ \cline{1-2}
MG-CIE         & CCRM        &                      &                      &                      &                          &                          &                        &                      \\ \hline
$\times$              & $\times$           & 75.58                & 83.56                & 81.04                & 79.36                    & 79.57                    & 59.38                  & 87.52                \\
\checkmark              & $\times$           & 74.92                & 87.52                & 83.54                & 80.91                    & 81.22                    & 63.01                  & 89.51                \\
$\times$              & \checkmark           & 73.27                & 88.28                & 83.54                & 80.32                    & 80.77                    & 62.89                  & 88.96                \\
\checkmark              & \checkmark           & \textbf{75.91}                & \textbf{89.49}                & \textbf{85.21}                & \textbf{82.40}                    & \textbf{82.70}                    & \textbf{66.07}                  & \textbf{90.52}                \\ \hline
\end{tabular}}
\end{table}

To further validate the effectiveness of the MG-CIE module, we integrated it into the comparative method MRIM, resulting in method MRIM+. As presented in Table~\ref{cr_men}, MRIM+ exhibits substantial improvements over MRIM across six metrics on the MEN dataset, with similar enhancements also seen in the derm7pt dataset (Table~\ref{cr_derm7pt}).
As depicted in Fig.~\ref{tsne_CCRM}, compared to MRIM, the t-SNE~\citep{van2008visualizing} visualization results of MRIM+ indicate tighter clustering of samples within the same class and increased distance between different classes, alleviating the modality gap. This suggests that MG-CIE has extracted more robust unimodal information and aids in exploring subsequent class-aware crossmodal relationships.

We also compared single-granularity CIE (denoted as SG-CIE) with multi-granularity CIE (MG-CIE). 
SG-CIE explore intermodal relationships only on the final outputs of encoders.
The comparison results for both datasets are presented in Table~\ref{cr_men} and Table~\ref{cr_derm7pt}, where Proposed- incorporates SG-CIE and Proposed uses MG-CIE.
In both datasets, Proposed demonstrates significant improvements over Proposed- across all metrics, highlighting the importance of relationship exploration and supplementary information extraction from multiple granularities.

\subsubsection{Effectiveness of the CCRM strategy}
The ablation studies of the CCRM strategy on the MEN dataset are presented in the first and third rows of Table~\ref{ab_men}. The incorporation of the CCRM strategy resulted in notable improvements across six metrics, particularly a $9.36\%$ increase in the ACC\_G2ninv metric and a $2.46\%$ increase in the ACC\_G2inv metric. 
Similarly, Table~\ref{ab_derm7pt} shows that the CCRM strategy also improved six metrics for the Derm7pt dataset, particularly a $4.72\%$ increase in the SPE metric and a $3.51\%$ increase in the AUPRC metric.
\begin{figure}
    \centering
    \includegraphics[width=\linewidth]{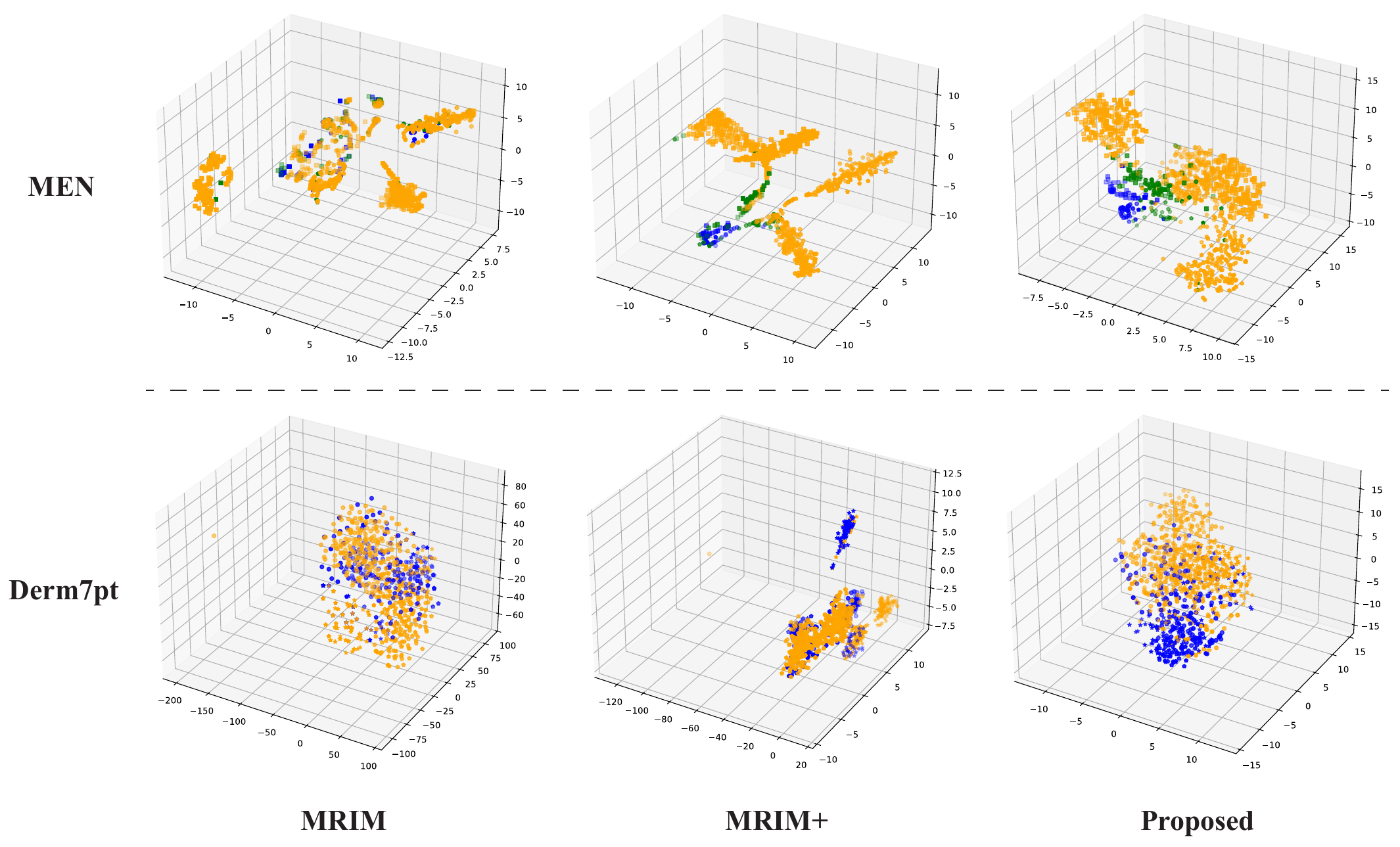}
    \caption{Display of t-SNE visualization results for proposed and comparison methods. In each sub-figure, different colors represent samples belonging to distinct classes, while different shapes denote samples from various modalities. The samples exhibiting different shapes but sharing the same color are more compact, while those of varying colors are more dispersed, suggesting that the model is more effective in reducing modality gap and extracting richer class-aware information.}
    \label{tsne_CCRM}
\end{figure}

We also visualize the crossmodal relationships of samples in the testing set for both datasets, as depicted in Fig.~\ref{tsne_CCRM}.
It illustrates that samples of the same class, indicated by the same color, are clustered together, even if they originate from different modalities, represented by different shapes. 
This observation suggests that our proposed CCRM strategy effectively extracts class-aware information from each modality, bridging the inherent modality gap.
Additionally, we compare the visualization results between MRIM+ and our proposed method. The increased compactness within the same class and the greater separation between different classes across both datasets further validate the effectiveness of our proposed CCRM strategy.

\begin{figure*}
    \centering
    \includegraphics[width=\linewidth]{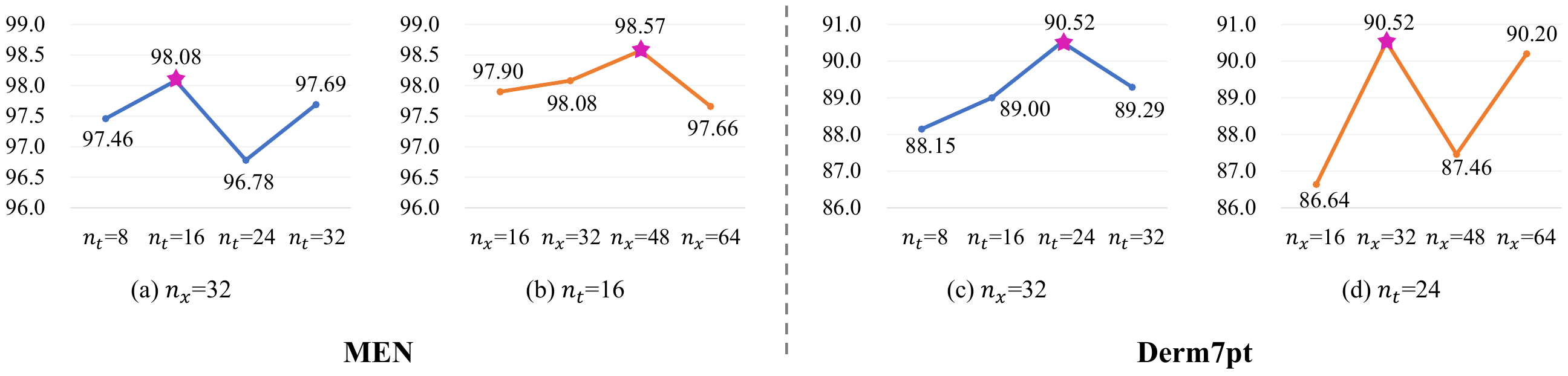}
    \caption{Display of the ablation study results about various predefined token numbers on both adopted datasets.}
    \label{tn_ablation}
\end{figure*}
\section{Discussion}
\label{discussion}

\subsection{The impact of predefined token mapping number}
As mentioned in Section~\ref{section_method}, the significant disparity in token mapping numbers between images and tabular data can lead to an information imbalance. Additionally, exploring the relationships among the numerous tokens necessitates substantial computational resources. Consequently, we mapped the token numbers of images and tabular data to predefined quantities. We performed ablation experiments with varying token numbers for both datasets, as illustrated in Fig.~\ref{tn_ablation}. Since images naturally possess more tokens than tabular data, we set the token range for images as $n_x\in\{16, 32, 48, 64\}$ and for tabular data as $n_t\in\{8, 16, 24, 32\}$. For the MEN dataset, we initially fixed $n_x$ at $32$ and identified optimal performance at $n_t=16$ (Fig.~\ref{tn_ablation} (a)). Subsequently, with $n_t$ held constant at $16$, we confirmed optimal performance at $n_x=48$ (Fig.~\ref{tn_ablation} (b)), thereby establishing the final values of $n_x=48$ and $n_t=16$. Following the same rationale, we determined $n_x=32$ and $n_t=24$ for the Derm7pt data.

\begin{figure}
    \centering
    \includegraphics[width=\linewidth]{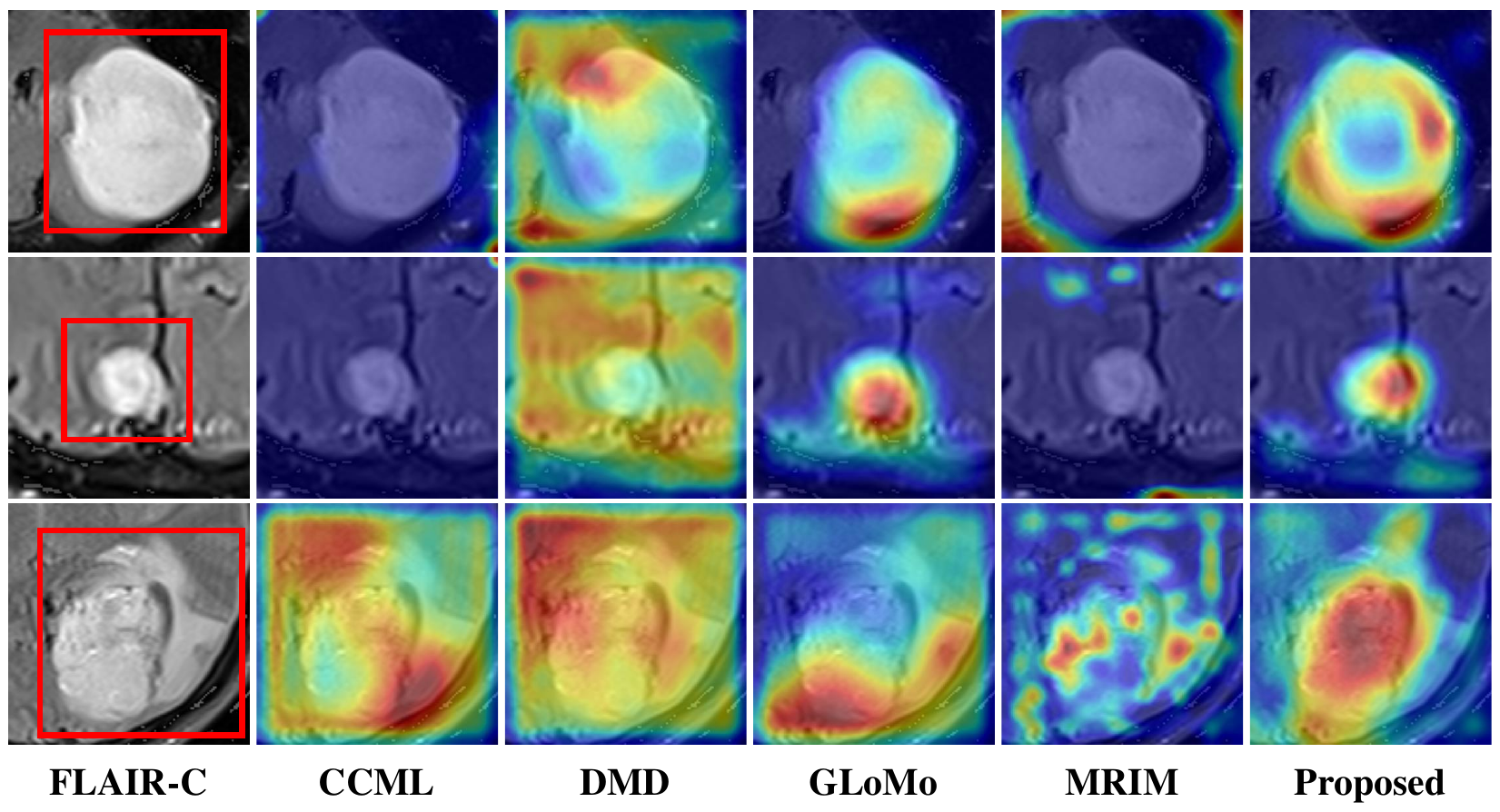}
    \caption{Activation map visualization results of the proposed and comparison methods for three cases in the MEN dataset using Grad-CAM. Original FLAIR-C images are displayed on the right, with the lesion areas highlighted by red boxes. In the Grad-CAM images, darker red regions signify the areas with the highest contribution to the prediction.}
    \label{grad_cam}
\end{figure}
\subsection{Visualization analysis}
The lesion area, which includes the tumor, its surrounding region, and edema, is critical for predicting meningioma grades in clinical research~\citep{hess2018brain,li2019presurgical,chen2023radiotherapy}. To enhance the interpretability,
we employed Grad-CAM~\citep{selvaraju2017grad} to visualize the activation maps of the proposed and comparison methods\footnote{We selected one method for each type of CML for comparison.} applied to FLAIR-C images across three cases in the MEN dataset, as illustrated in Fig~\ref{grad_cam}. This figure demonstrates that the proposed method emphasizes the lesion area, indicated by red boxes in the FLAIR-C images\footnote{Both the tumor and edema areas are highlighted in the FLAIR-C image; we utilize this modality for visualization.}, underscoring its effectiveness in identifying critical regions relevant to the prediction of meningioma grade.

\begin{figure}
    \centering
    \includegraphics[width=0.9\linewidth]{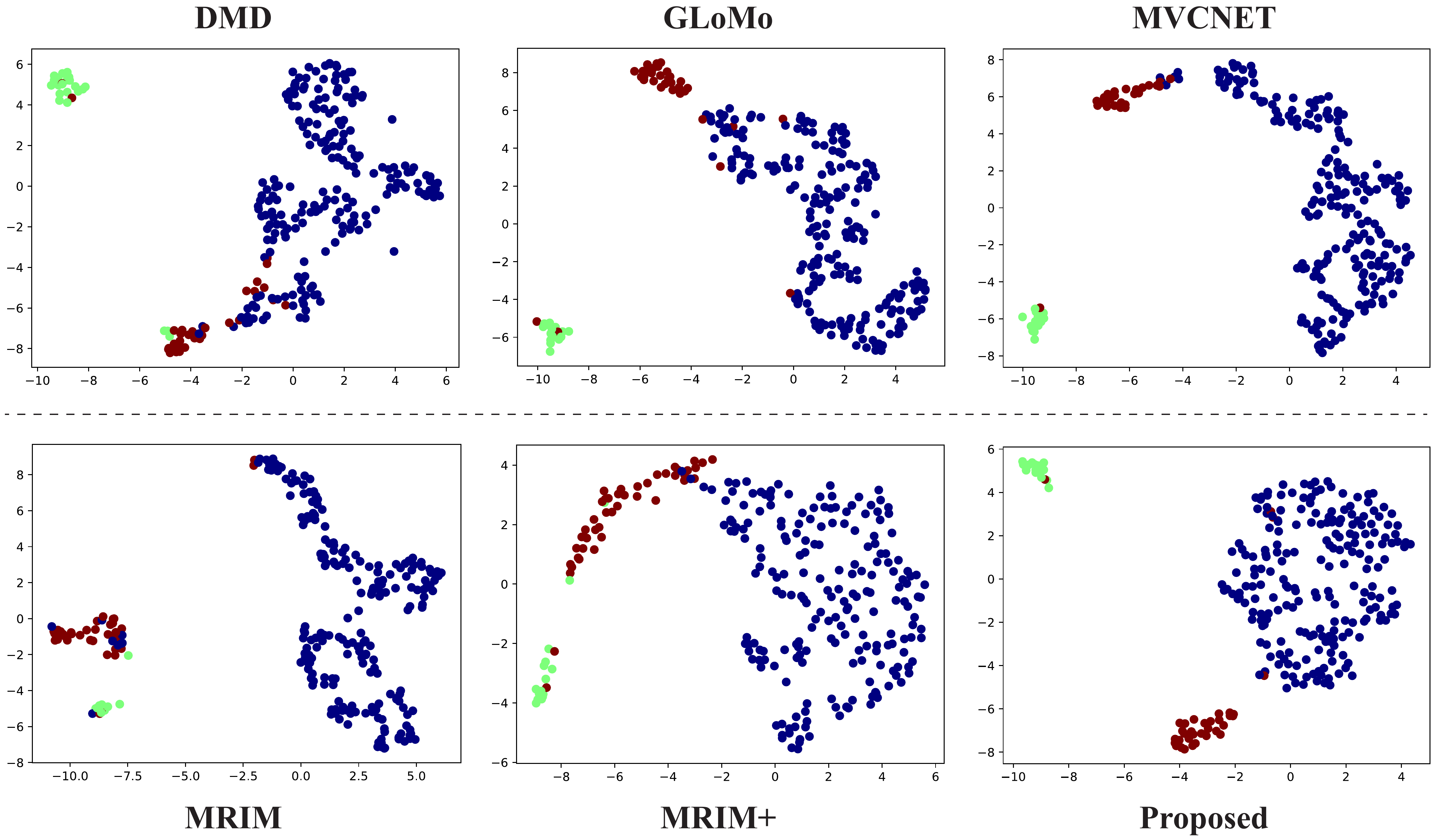}
    \caption{Display of MDA visualization results for proposed and comparison methods. In each sub-figure, different colors represent samples belonging to distinct classes.}
    \label{mda}
\end{figure}

To validate the discrimination of the fused crossmodal features, we used the Manifold Discovery and Analysis (MDA)~\citep{islam2023revealing} algorithm to visualize the feature space distribution for the proposed and comparison methods\footnote{We did not compare the uncertainty-based methods because these methods involve decision-level fusion and do not generate fused crossmodal features.} on the MEN dataset, as illustrated in Fig.~\ref{mda}. Compared to the SOTA methods, the visualization results for the proposed method exhibit clear boundaries between different class samples and fewer misclassifications, indicating that the crossmodal features derived from our proposed method are more discriminative, owing to the MG-CIE module and CCRM strategy.

\subsection{Computational complexity analysis}
We analyzed the computational complexity of the comparison methods, with results shown in Table~\ref{cr_men} for the MEN dataset and Table~\ref{cr_derm7pt} for the Derm7pt dataset. As mentioned in Section~\ref{section_method}, we configured the encoder of the comparison methods to be identical to ours; however, the comparison methods did not explore the multi-granularity relationships across modalities. To ensure fairness, we compared the Proposed- method (with the SG-CIE module) against the comparison methods. Specifically, the Proposed- method utilized relatively fewer parameters and GFLOPs while achieving notable improvements over the comparison methods across multiple metrics on both adopted datasets.

It is observed that the proposed method incurs slightly higher costs in terms of parameters and GFLOPs compared to the Proposed- method. This increase is primarily attributed to the $Conv1d$ operations employed to map the crossmodal tokens to a predefined value across multiple granularities. Specifically, there are 16 $Conv1d$ operations for four input modalities.
Furthermore, we compared our proposed method with MRIM+, which incorporates the MG-CIE module into the MRIM method and has similar parameters and GFLOPs; however, our method shows significant improvements.

\begin{figure}
    \centering
    \includegraphics[width=0.8\linewidth]{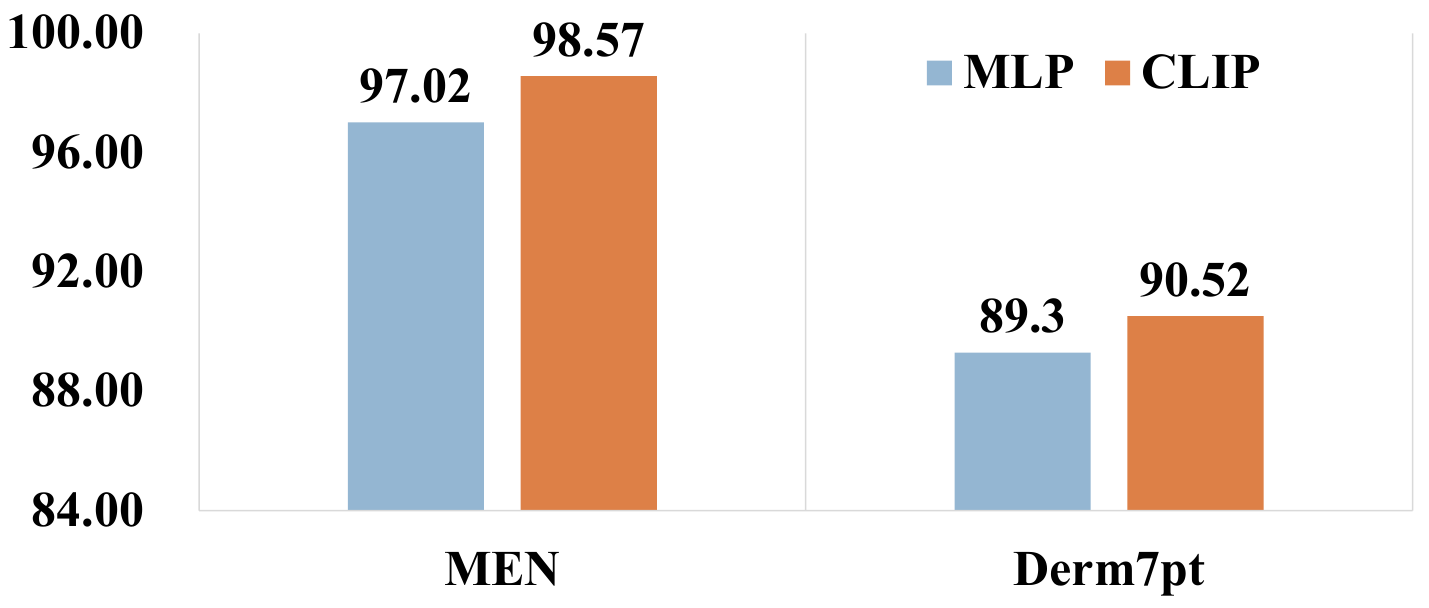}
    \caption{The comparison results of the AUC metric for CLIP embedding and MLP embedding of tabular data on the MEN and Derm7pt datasets.}
    \label{comparison_clinical_embed}
\end{figure}
\subsection{Comparison of tabular data embedding methods}
We conducted a comparison of embeddings for tabular data across both adopted datasets, including the pretrained CLIP and the traditional multi-layer perceptron (MLP) method, as illustrated in Fig.~\ref{comparison_clinical_embed}. This figure demonstrates that the CLIP embedding outperforms the MLP embedding by $1.55\%$ and $1.22\%$ for the MEN and Derm7pt datasets, respectively. These results indicate that the pretrained CLIP is more effective in extracting relevant information from tabular data compared to the MLP.

\begin{figure}
    \centering
    \includegraphics[width=0.85\linewidth]{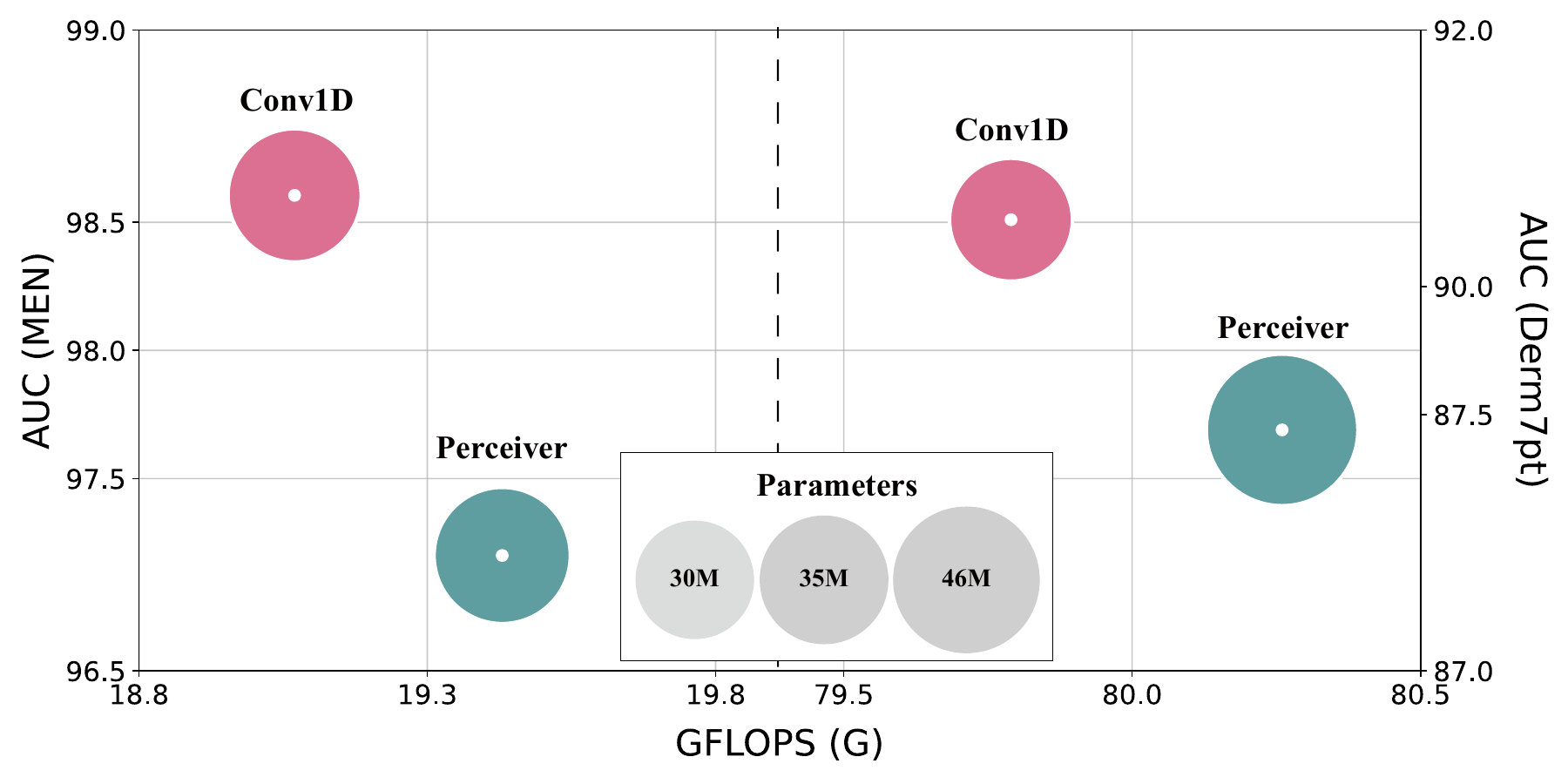}
    \caption{The comparison results for different token mapping operations, including $Conv1D$ and $Perceiver$, on the MEN (left part) and Derm7pt (right part) datasets.}
    \label{vis_bc}
\end{figure}
\subsection{Comparison of different token mapping methods}
We compared different token mapping operations, including $Conv1D$ and $Perceiver$~\citep{jaegle2021perceiver}, as shown in Fig.~\ref{vis_bc}. In both datasets, $Conv1D$ outperformed $Perceiver$ in AUC metrics while requiring fewer parameters and GFLOPS. This indicates that $Conv1D$ is a more efficient and effective approach for token mapping in the proposed method, reducing model complexity while maintaining high performance.

\begin{figure}
    \centering
    \includegraphics[width=0.85\linewidth]{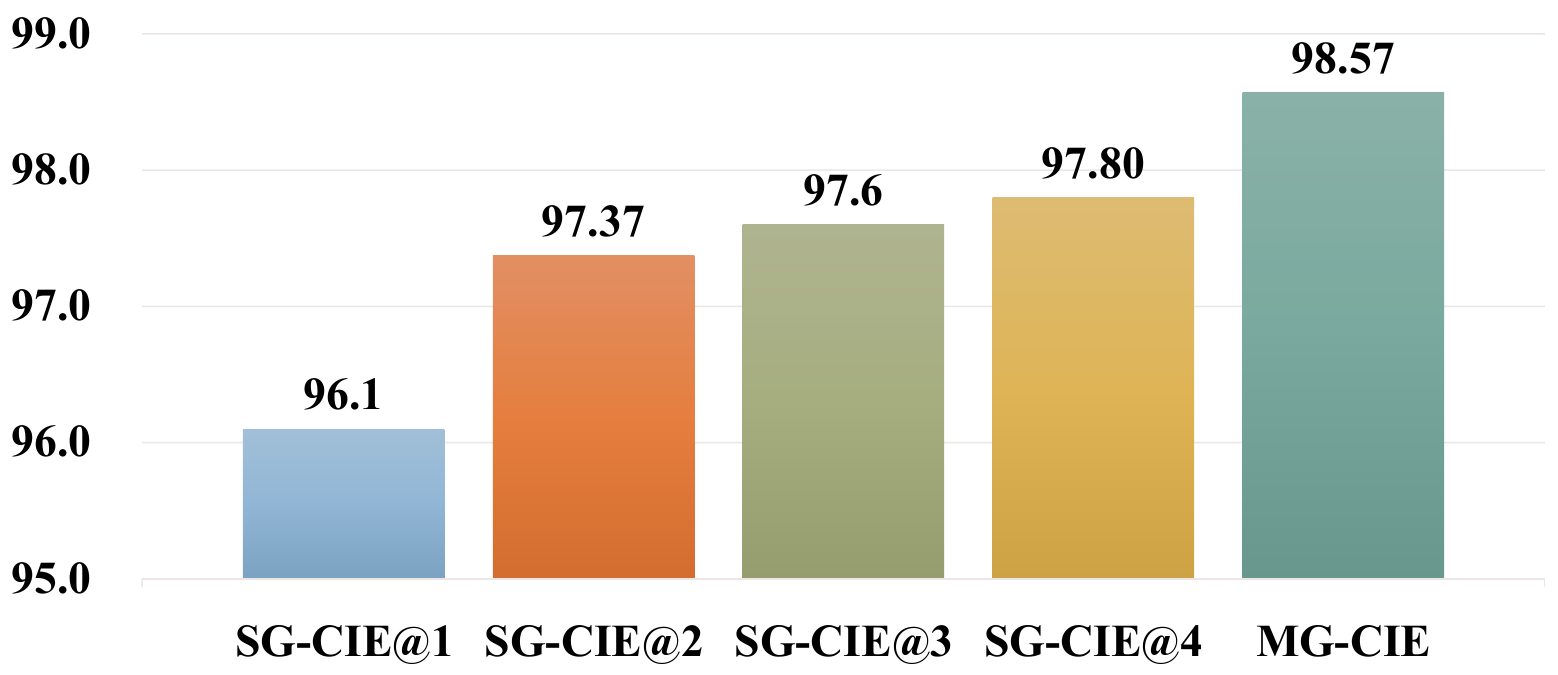}
    \caption{The ablation results for the AUC metric of SG-CIE at various stages on the MEN dataset. The notation ``SG-CIE@i'' denotes SG-CIE at the $i$th stage of the encoders.}
    \label{as_singlelevel}
\end{figure}
\subsection{Ablation of SG-CIE at various stages}
We compared the results of SG-CIE at various stages on the MEN dataset, as illustrated in Fig.~\ref{as_singlelevel}. The figure indicates that the results improve as the stage of SG-CIE increases, suggesting that a deeper encoder stage yields better-extracted features. Furthermore, the performance of MG-CIE surpasses that of all SG-CIE at each stage, thereby demonstrating the effectiveness of exploring relationships between multi-granularity features across modalities.

\section{Conclusion and future works}
In this paper, we propose a novel coarse-to-fine crossmodal learning (CFCML) framework that progressively reduce the modality gap between multimodal images and tabular data by thoroughly exploring intermodal relationships. We first design MG-CIE module to preliminary reduce the modality gap and enhance unimodal feature by exploring the interactions between the multi-granularity image features and tabular information. To further reduce the modality gap and extract class-aware information, we introduce the CCRM strategy, which establishes the unimodal and crossmodal prototypes and develops hierarchical anchor-based CL strategies. These strategies aim to cluster modalities belonging to the same class while pushing those from different classes apart, thereby bridging the boundaries across modalities. Experimental results and visualization analysis demonstrated the superiority of our proposed method over other SOTA CML methods.

Despite the proposed CFCML method demonstrating excellent performance, it has several limitations. First, the exploration of interactions from multi-granularity crossmodal features in the MG-CIE leads to an increase in computational complexity. Second, manual adjustment of the token mapping number across different datasets is necessary, which requires a considerable amount of time to determine the optimal value. Future work will focus on developing fully automated and parameter-efficient methods.

\printcredits

\bibliographystyle{cas-model2-names}

\bibliography{ref}



\end{document}